# May We Have Your Attention:
# Analysis of a Selective Attention Task

**Eldan Goldenberg[1], Jacob Garcowski[1] and Randall D. Beer[1,2]**
[1]Dept. of Electrical Engineering and Computer Science
[2]Dept. of Biology
Case Western Reserve University
Cleveland, OH 44106
{exg39, jrg13, rxb9}@cwru.edu

## Abstract

In this paper we present a deeper analysis than has previously been carried out of a selective attention problem, and the evolution of continuous-time recurrent neural networks to solve it. We show that the task has a rich structure, and agents must solve a variety of subproblems to perform well. We consider the relationship between the complexity of an agent and the ease with which it can evolve behavior that generalizes well across subproblems, and demonstrate a shaping protocol that improves generalization.

## 1. Introduction

Within the adaptive behavior community, it is widely believed that situated, embodied and dynamical approaches have significant implications for cognitive science (Miller, 1994; Clark, 1997; Harvey, 2000). In order to realize this potential, however, we must create model agents capable of *minimally cognitive behavior*, the simplest behavior that raises issues of genuine cognitive interest. Toward this end, we have previously evolved dynamical "nervous systems" for a variety of minimally cognitive behaviors, including categorical perception, short-term memory and selective attention (Beer, 1996; Slocum et al., 2000).

Our primary motivation for these studies has been to investigate the mechanisms by which the evolved agents operate. For example, a detailed dynamical analysis of an evolved agent capable of visually-guided object discrimination has recently been carried out (Beer, in press). This analysis involved characterizing (1) the dynamics of the entire evolved brain-body-environment system, (2) the dynamics of the agent and environment subsystems and the interactions that give rise to the observed behavior of the coupled system, and (3) the neuronal properties underlying the agent dynamics.

In attempting to apply these techniques to more complex behaviors, we have found it necessary to first understand in detail the structure of the minimally cognitive tasks that we set. While simple tasks may have only one or a small set of obvious best solutions, more complex tasks may involve tradeoffs that interact in complicated ways. These tradeoffs must be understood before a detailed dynamical analysis of agents performing such tasks can be undertaken.

In this paper, we perform a task analysis of the selective attention problem. First, we present a largely unsuccessful attempt to quantify the difficulty of this task. We then demonstrate that this attempt fails because the task has considerably more structure than is initially apparent, containing many different subproblems on which different evolved agents specialize in different ways. Next, we present the beginnings of a taxonomy of these subproblems and their particular difficulties, and show that the overall success of different agents can be understood in these terms. Finally, we demonstrate that, in random trials, these different subproblems occur with frequencies that sometimes differ by orders of magnitude. By manipulating these frequencies, we can put selective pressure on the evolution of agents that perform better on the harder subproblems, or that generalize better across the full range of subproblems.

## 2. Methods

Agents are trained to handle a selective attention task (Slocum et.al., 2000; Downey 2000). An agent, with limited sensory capability, is required to catch two falling objects. The agent is represented by a circle, and is capable of moving horizontally. The falling objects, also circles, have separate constant velocity, and are constrained such that the agent should be capable of catching both.

An agent is a circle of diameter 30, having 9 proximity sensors of range 205 evenly distributed over a visual angle of $\pi/6$. The world the agent exists in is 400 units wide. The agent can not move outside of this world, however, it also has no way to know that it has been stopped at the edge. The agent's horizontal velocity is proportional to the sum of two opposing motor neurons' output. The constant of proportionality is 5, thus, the maximum velocity of the agent in either direction is 5. The falling objects have diameter 26.

The first object's vertical velocity is in the range [3,4], while the second object's vertical velocity is in the range [1,2]. Both objects horizontal velocities are in the range [–2, 2]. The objects are constrained to start within the agent's field of view, and travel in such a way as it is always possible for the agent to catch both. The faster object will always land first. The second object will land so that it can be reached by the agent traveling at $5\alpha$, where $\alpha$ is a scaling factor, set to 0.7, to ensure the agent would be capable of catching both objects. When an object lands, it is removed from the simulation, so that it does not confuse the agent's visual sensors. See Figure 1 below.

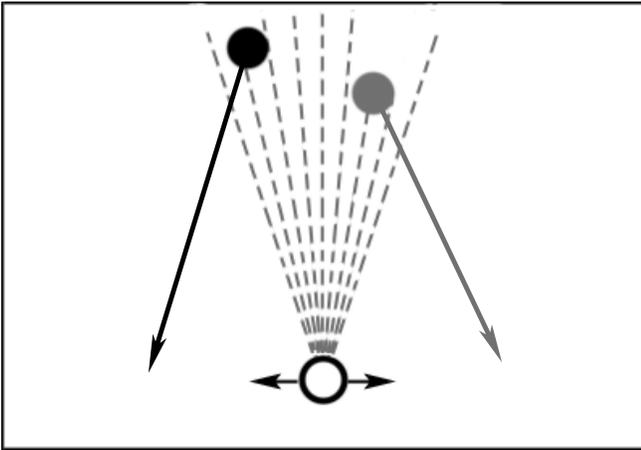

Figure 1: The agent can move horizontally in order to catch the 2 objects which fall from above.

An agent's behavior is controlled by a continuous-time recurrent neural network (CTRNN) with the following state equation:

$$\tau_i \dot{y}_i = -y_i + \sum_{j=1}^{N} w_{ji}\, \sigma(g_j(y_j + \theta_j)) + I_i \quad i = 1, \ldots, N$$

Where $y$ is the state of each neuron, $\tau$ is its time constant, $w_{ij}$ is the strength of the connection from the $j^{th}$ to the $i^{th}$ neuron, $g$ is a gain, $\theta$ is a bias term, $\sigma(x)=1/(1 + e^{-x})$ is the standard logistic activation function, and $I$ represents an external input (zero on all but sensory neurons). A forward Euler method with a step size of 0.1 was used to integrate the CTRNN.

The CTRNN is constrained to be bilaterally symmetric, with 9 sensory neurons, a set even number of interneurons, and 2 motor neurons. Sensory neurons only receive input from the environment, but they connect to all interneurons and motor neurons. The interneurons and motor neurons are fully interconnected (every interneuron or motor neuron is connected to every other interneuron or motor neuron). Agents are constrained to bilateral symmetry both to simplify the problem and to ensure that an agent will act the same in mirror situations. All sensory neurons share the same gain and bias, interneuron and motor neuron biases are constrained to the range of [–5,5], and motor neuron gains are fixed at 5. All gains are positive, and all time constants are greater than 1.

The parameters of agents were evolved using a real-valued genetic algorithm (Mitchell, 1996). An individual agent was represented by a real-value vector of length M. Initially a population of 100 individuals was created by setting their evolvable parameters to random numbers in the range [–1, 1]. The top 2 individuals were copied into the next generation automatically. The remaining individuals were created by mutation. A linear-rank based method was used to select individuals for mutation. A selected parent was mutated by adding a vector whose direction was randomly distributed on an M-dimensional hypersphere and whose magnitude was a Gaussian random variable with mean 0 and variance $\sigma^2$, of 1. Agents were trained over 9000 generations.

An agent's performance on any individual trial was (200 – $p$), where $p$ is the error, measured as the sum of the absolute distances between the center of the agent and the center of the falling objects when they land. Therefore, the maximum performance of an agent on any trial is 200. Agents are ranked based upon their average performance for a set of trials. For purposes of reporting, agent performance will be given as a percentage of the maximum possible.

Agents were trained using a shaping scheme. As evolution continues, the agents were presented with more difficult problems. Each agent was trained on a set of 30 trials. Over time, the trials that the agent performs best on were replaced with others. Whenever the best agent performed better than a threshold T, or 600 generations passed since the last trial change, the trial that it performed best at was replaced with a new trial. The first 5 of these new trials were hand selected. Subsequently, new trials were generated by selecting the first random trial that the agent performs worse than 170 on. T = 198 – $n$/14 – $gen$/2500, where $n$ is the number of additional trials added, and $gen$ is the current generation. For work using similar incremental shaping schemes to guide evolution or learning towards flexible behavior, see (Dorigo & Colombetti, 1994; Saksida et.al., 1997).

## 3. Task Difficulty

How difficult is this selective attention task? A more difficult task should provide for a more interesting minimally cognitive behavior study. A good way to show the difficulty of a task is to consider how complex an agent is needed to solve it. For our task, we trained agents with differing numbers of interneurons. 16 genetic algorithm runs were done for each number of interneurons.

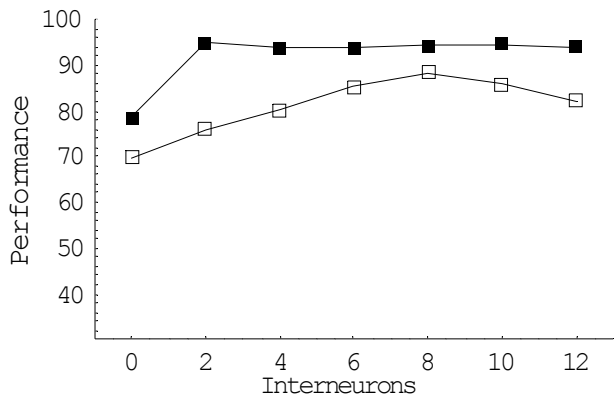

Figure 2: The best and average performance for each set of interneurons. Best performance is solid boxes, average is outlined boxes.

Figure 2 shows an unexpected result: agents with low numbers of interneurons are still capable of performing the task as well as higher numbers of interneurons. The best 2 interneuron agent is able to perform as well as all of the other agents. The best performance only drops when zero interneuron agents (purely reactive agents, with feed forward connections from the sensor neurons to the motor neurons) are considered. An agent with two interneurons should not be able to hold as much internal state as agents with 8, 10, or 12 interneurons, and yet is still capable of performing as well. This seems to go directly against our beliefs that this task requires both short-term memory and selective attention, cognitive abilities that should require more internal state. Previous work on this problem (Downey, 2000) suggested that the ideal number of interneurons would be 10, and yet a 2-interneuron agent can perform as well as a 10 interneuron agent.

The mean values in Figure 2 show a more expected shape. These values suggest that while a 2-interneuron agent is capable of solving the problem, training it to do so is more difficult.

At first, the surprisingly small difference in performance between the best agent of each type would seem to imply that the task is simpler than had previously been assumed (Downey, 2000). We contend that it is misleading to treat this task as uniformly complex. Closer analysis shows that the complexity of the task, and of the agent architecture required to accomplish it, varies considerably from one trial to another. The interaction between the random task parameters makes some trials entirely trivial, but others are more difficult for the agents.

## 4. A Taxonomy of Subproblems

Detailed analysis of the task parameters and their effects shows that there is a wide variety of subtasks, ranging from the entirely trivial to very difficult cases on which only a small proportion of agents perform well. Previous work (Downey, 2000) considered four subgroups of trials, and found significant variance between the agents' performance across these subgroups. We are incrementally refining this classification, to which end we divided a sample of 100,000 randomly generated trials into 24 mutually exclusive categories according to salient features of the setup.

Comparisons of several agents' performance across these trial categories showed that many could be meaningfully grouped together, and that it was more useful to base these groupings on the demands they make of the agent, or the strategies an agent could use to perform well on them. This produces overlapping groups, which we describe below. After eliminating the easiest trial types, we consider those that require some short-term memory and those that set a selective attention task more difficult than simply ignoring the more distant object from the start of the trial, before looking at some special cases.

Subsequent analysis of the individual agents' performance will focus on their relative strengths and weaknesses, from the perspective of the problems they have to solve in particular trial groups. Rather than presenting a full taxonomy, this section will set out the main groupings that will be used to illustrate the evolved agents' strengths and weaknesses.

### 4.1 Trivial trial types

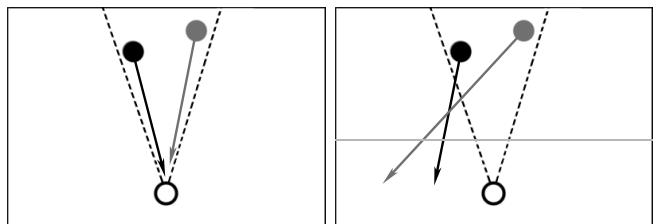

Figure 3: Trivial trial types: an instance for which the agent need not move at all (left) and one for which it need only move reactively (right). *In all figures in this section the agent is shown as an empty circle, its field of vision is between the dashed lines, the objects are shown as filled circles with their trajectories as arrows, and the darker object (which starts to the left) lands first. Here a horizontal line indicates the second object's height when the first lands.*

Some trials make very few demands of an agent, and therefore provide little information when analyzed. An obvious example of this is that it is possible for both objects to land directly on the agent's starting position, in which case the agent will get a perfect score by simply not moving at all. This caricature example is rather rare, accounting for only 5.2% of trials, but there is a more common subgroup of trials which can be dismissed as trivial because they only require purely reactive behavior of the agent. Any trial in which the object that starts closer to the agent is the first to land, and the second object is within the agent's field of vision when the first is caught, is trivial in this sense. Altogether, such trials account for 31% of the total. An example of each of these is shown in Figure 3 above.

## 4.2 Selective attention problems

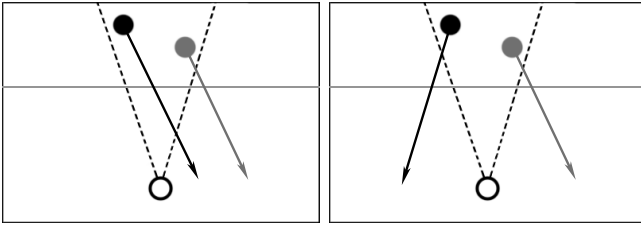

Figure 4: "delayed decision" (left) and "unseen passing" trials (right). *Here the horizontal lines indicate the height at which the darker object overtakes the lighter one.*

Selective attention becomes a more difficult problem when the object that will land first starts further away from the agent than the other. Two examples of this are shown in Figure 4 above. We will refer to trials in which the object that starts further away lands first as "delayed decision" trials, because the agent can not correctly choose which object to catch first until it has observed them moving. These instances make up 48% of all randomly generated trials[1].

A more difficult subset of the "delayed decision" trials occurs when the faster object passes the slower one at a large enough horizontal distance that the agent can not see both. In these cases, the correct choice of which object to follow can no longer be made based on distance information alone; the objects' relative speeds must be considered as well. We call such instances "unseen passing" trials, and they account for only 0.099% of the total.

## 4.3 Short-term memory problems

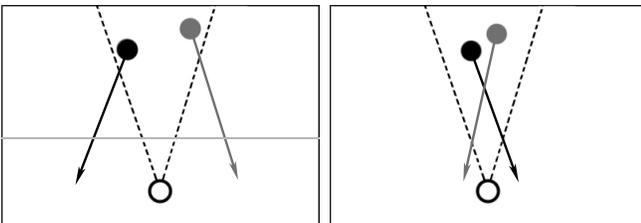

Figure 5: "object permanence" (left) and "overlapping objects" (right) trials. *Here the horizontal line indicates the height of the second object when the first lands.*

Another class of trials requires the agent to store information so that it can find an object again after having to lose sight of it. We consider two such situations, as illustrated by Figure 5 above.

If the second object is not within the agent's field of vision when the first is caught, then the agent must move towards the second object without being able to see it. This requires both that the agent stores information about the unseen object, and that it switches its attention to this internal information in response to catching the first object. These "object permanence" trials account for 39% of the total.

---

[1] Such instances account for less than 50% of all trials because randomly generated trials in which the two objects will land too far apart for the agent to catch both are rejected. This rejection criterion affects a slightly higher proportion of parameter sets in which the object that starts higher also moves faster.

An additional difficulty is presented when the two objects cross horizontally, or start with a horizontal overlap. The poor sensory resolution of the agent means that as long as a group of neighboring rays are all broken, it will not be clear whether they are being broken by one large object or multiple smaller ones, and one object can be entirely occluded by the other. In such cases, which account for 58.6% of the total, it is possible for the agent to follow the correct object at the start but be fooled by the overlap, and switch its attention as a result.

## 4.4 'Cheat' heuristics and their limitations

For each subset of trials, it is possible for an agent to catch both objects by the using one of various simple 'cheat' heuristics. However, each such heuristic applied uniformly will only solve a small proportion of all trials; reliable performance requires a more flexible set of responses from the agent. For instance, an agent which always follows the object that starts further away will be successful on the delayed decision trials, but its overall performance will be poor because it will fail to catch the first object on the more straightforward trials.

The short-term memory requirement can also be circumvented for a subset of trials by the agent automatically moving in a pre-set direction after catching the first object if the second is not within sight, and switching to reactive behavior once it has moved far enough that it can see the second object again. However, any such behavior will guarantee failure to catch the second object on a proportion of trials (as shown in Figure 8 below). The most successful pre-set heuristic would be to continue in the same direction after the first catch, which will succeed in 78% of "object permanence" trials (making up 30% of all trials), but still guarantees failure in the remaining "object permanence" trials, which account for 8.4% of all random trials.

## 4.5 Summary

The taxonomy presented above is not a comprehensive classification of all the subtleties of this task, but a step towards this goal. The variety of different subtasks, different combinations of which are involved in each random trial, ensures that agents must have some degree of behavioral flexibility to perform well across all trial types. The following section will illustrate the interaction between individual agents and different trial types.

# 5. Individual Variation

The agents we have evolved show a wide variety of behavior patterns, and relative strengths and weaknesses. In this section we present some illustrative examples of the range of agents we have evolved, and how they are differentially affected by the complexities of the task.

We use two different performance metrics to compare the performance of agents on various subsets of trials. The fitness measure described in Section 2 above will be used, but it is supplemented with a binary measure of whether each object is caught by the agent on each given trial, because this can make analysis clearer. A catch is defined as any collision between the agent and the object, without necessarily having to be exactly aligned, and we report the percentage of trials in which the agent catches each or both objects. The agents were tested on the same batch of 100,000 randomly generated trials described in Section 4 above, and their performance was recorded for various subsets of these trials defined by the combinations of parameters.

## 5.1 A blind spot for the best agent

It is instructive to begin the analysis with the best performing agent that we have generated to date. This agent has 10 interneurons, an overall fitness score of 97%, and catches both objects in 99% of all trials. It performs reliably well across almost all trial categories, catching both objects with 100% reliability in many, and with better than 97% reliability in all but one. However, its performance in the "unseen passing" situation is strikingly poor: earning a fitness score of 27%, and it only catches both objects in 20% of these trials.

In fact, closer analysis shows that this agent almost always (95% of trials) succeeds in catching the second object in these cases; it is the first object that it has trouble with (only catching it in 25% of trials). An example of its typical behavior is illustrated by figure 6 below. It seems that this agent has evolved a strategy of choosing the closer object when it is no longer possible to keep both within sight. In most trials (all but the "unseen passing" type described in Section 4.2 above, which only accounts for 0.099% of the total) this strategy will lead to a correct choice of the first object. It does fail consistently on one subgroup of trials, but because this subgroup is so rare it still allows the agent to score impressively well overall.

This striking weakness in an otherwise highly performing agent shows that the task is not uniformly complex, but rather that special cases provide particular challenges for the agent. Conversely, the trial type on which a particular agent performs worst will not necessarily be the hardest type for all other agents, as the next subsection will show.

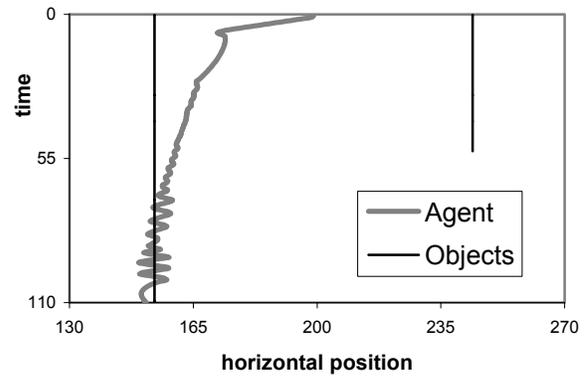

Figure 6: The best agent usually fails to catch the first object in the "unseen passing" situation. *The first object's trace ends when it reaches the agent's level.*

## 5.2 "Unseen passing" trials are tractable

There are also agents for which the "unseen passing" trials present little or no handicap, but which have weaknesses on other trial types. The best performance on this subgroup of trials was from an agent with only 4 interneurons, which scored almost as well on this subset as on average. The overall fitness for this agent is 91%, and it catches both objects in 92% of all trials; clearly less reliable than the agent described in 5.1 above. However, in the "unseen passing" situation it catches both objects in 89% of trials.

There are other subsets of trials which give this agent more trouble than the "unseen passing" ones. Its worst performance is for a subset that we were not expecting to provide particular difficulties: where the first object starts directly above the agent, and both objects fall diagonally in the same direction, without crossing in the horizontal plane, as illustrated in Figure 7 below. These trials account for 0.74% of the total, and this agent manages to catch both objects in only 76% of such trials.

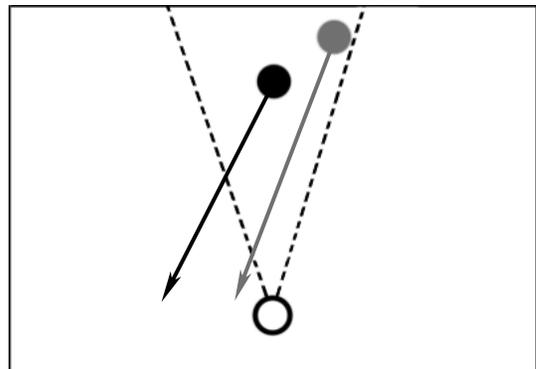

Figure 7: The trial setup on which the agent described in Section 5.2 performs worst.

## 5.3 Fixed action patterns

Both of the agents described above perform well enough in general that they do seem to be responding to the task in a generalized manner. There are others that show evidence of using a much simpler approach. Another 4-interneuron agent provides a particularly striking example of this.

This agent's overall performance is not impressive: it has a fitness of only 59%, and catches both objects in 67% of all trials. However, there is a subset of nontrivial trials for which its performance is far better than this. For trials which can be solved by rigidly applying the heuristic of "follow the closer object, and then continue to move in the same direction to find the other after catching it" (as shown in Figure 8 (left) below) this agent catches both objects in 97% of trials. By contrast, it only catches both objects in 46% of trials for which it would have to reverse its movement after the first catch (shown in Figure 8 (right)).

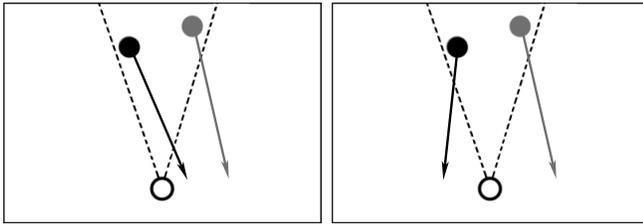

Figure 8: The types of trial soluble by two different fixed action patterns: Continuing in the same direction after catching the first object (left) or reversing direction (right)

Breaking down the data further shows that this agent catches the first object reliably in most trial types (99.9% of those shown in Figure 8 (left), 93% of those shown in Figure 8 (right), and 97% of all trials), so the variation in its performance comes almost entirely from what it does *after* the first catch. It catches the second object with 97% reliability in the "continue in the same direction after the first catch" cases, as compared with only 53% in the "reverse after the first catch" trials, and 70% overall.

A similar pattern, though less pronounced, is shown by 4 other 4-interneuron agents out of the 16 we have evolved, so while it does not account for the majority of results it is clearly not a unique instance. Our speculative explanation is that this is a comparatively easy strategy to evolve within the limitations of the 4-interneuron architecture, and because the trials on which this strategy fails account for a relatively small proportion of the total, the poor performance on these does not drag down overall fitness enough for these agents to always die out in evolution. More analysis needs to be done before this can be stated authoritatively.

## 5.4 What can be done with 2 interneurons

In general, it does appear harder to evolve agents that perform well with smaller numbers of interneurons. Agents with fewer interneurons also tend to perform more inconsistently across different trial types, indicating that they are more likely to rely on simple 'tricks' to solve the majority of subproblems. However, the best performing 2-interneuron agent performs surprisingly well across all trial types.

This agent has an overall fitness of 94% (as compared with the best agent of all, which was described in 5.1 and scores 97%), and catches both objects in 97% of all trials (as compared to 99% for the best agent). Its weakest performance is on the same "unseen passing" trials that stymie the best agent, but it is actually far less handicapped by this subset, earning a fitness score of 74% and catching both objects for 68% of such instances (as compared to only 20% for the otherwise superior agent described in 5.1).

One possible explanation for this agent's high overall performance is that it has developed a sort of toolkit of simple solutions to specific subproblems, and is able to select the right one in most cases. However, the standard deviation for this agent's reliability across all of the trial subtypes we have analyzed is 6.8 (as compared to 27 for the particularly inconsistent agent described in 5.3 above) indicating overall flexibility of behavior. Furthermore, it is hard to see how an agent with only two interneurons–and therefore only two pathways from visual input to motor output–could have a comprehensive repertoire of 'programmed' behaviors.

All of these data point to this agent having evolved a general strategy that is robust to the various different types of problem presented by different trial parameters, in spite of having such a simple control architecture. However, this is not true of any other 2-interneuron agents in the 16 we have evolved. It seems to be the case that while more interneurons are not necessary to perform well on this task, they make evolving reliable solutions considerably easier. This is borne out by the difference between the mean fitness of agents with 2 interneurons and those with 10, as shown in Figure 2 above; while this good 2-interneuron agent is an outlier, the best 8 or 10-interneuron agents are not far removed from the average performers in their class. This is a finding that needs more analysis and a larger number of evolution runs than time has permitted at this point.

## 5.5 Summary

The examples discussed above show that the task we are analyzing has a richly varied set of subtasks, which make different demands of our evolved agents. Because they are categorically different subtasks, rather than degrees of difficulty along one dimension, agents can develop idiosyncratic patterns of relative strengths and weaknesses, with a subtask that stymies one agent providing no trouble at all for another. The selective weaknesses of generally high performing agents can be at least partly explained by the rarity 'in the wild' of the trial types on which they fail; they have not evolved an appropriate response to situations that were either never or very rarely encountered during evolu-tion. Section 6 below presents an attempt at testing this hypothesis.

The consistently high performance of the best 2-interneuron agent proves that 2 interneurons are sufficient to develop a generalized response to the task, contrary to our previous expectations. However, the exceptional nature of this agent, contrasted with the high average performance of 10-interneuron agents suggests that 10-interneuron architectures are far more likely to evolve good solutions. It may be that producing such a range of behaviors with such a simple architecture requires a very precise, and therefore brittle, combination of parameters, though this is a tentative explanation that needs more work to support it.

## 6. Improving Generalization

By modifying the trials an agent sees during training, we can increase performance on any subproblem, hopefully allowing for greater generalization.

An agent is exposed to a surprisingly small number of trials throughout its evolution. An agent will always be evaluated on 30 trials. New trials are added by replacing the trial that the agent does best on. During training, an agent will be introduced to a new trial when it is performing around 99%, or 600 generations after the last new trial was introduced. The performance limit for introducing a new trial is prohibitively high, therefore, it can be assumed that an agent will be introduced to a new trial every 600 generations, so only 14 trials will be added during the 9000 generations of training[2]. The first 35 trials are hand selected to encourage a valid solution to be formed, the remainder are selected from a group of 30 randomly generated trials, where the first one found to cause the current top-performing agent to fail (score below 85%) is added. Informal observation says that it is unusual for more than 30 trials to be generated before the agent fails on one. Thus, a reasonable upper bound for the number of trials an agent will encounter during training is 305 (including the initial 35 trials).

There are a large variety of subproblems, and many are quite rare. An example is the "unseen passing" subproblem, which only occurs 0.099% of the time. With this extreme rarity, the expected number of "unseen passing" trails that an agent will see during training is less than 0.2673.

By changing the trials an agent is exposed to during training, we should be able to increase performance in different subproblems. To test this hypothesis, the shaping schedule was modified to include one "unseen passing" problem. By introducing one "unseen passing" problem early in training we give agents much longer to train on the "unseen passing" subproblem. If an agent was to encounter an "unseen passing" trial during standard shaping, it would most likely be in a later generation, giving the genetic algorithm less time to evolve good solutions. Figure 9 below shows the average performance on the "unseen passing" subproblem for agents trained normally and those trained with the modified shaping. Due to time constraints, only 4 agents could be evolved for each number of interneurons on the modified shaping. 16 agents were evolved for each number of interneuron using standard shaping. The average score for agents with modified shaping is higher than standard agents, showing that an agent's performance on a subtask can be improved by modifying the shaping. While the best performing agents were evolved with the standard shaping, we believe this is an artifact of having 4 times more standard shaping agents.

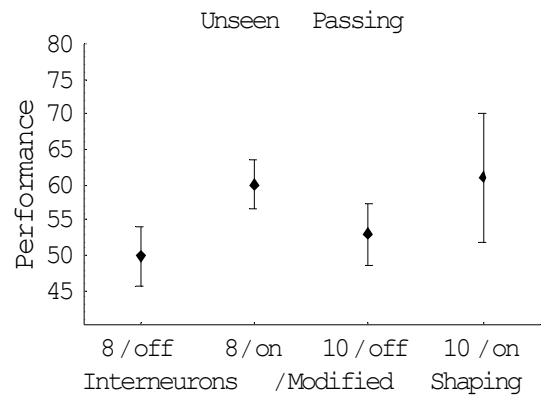

Figure 9: Performance on "unseen passing" subproblem with and without shaping modified to include an "unseen passing" trial. Error bars represent the standard error of the averages.

## 7. Conclusion

In this paper we have revisited a selective attention task which has previously been considered (Slocum et al., 2000; Downey, 2000) and analyzed the problem space in greater depth. Empirical observation of a selection of agents evolved for this task has shown that their performance does not vary monotonically across parameter sets. Instead, there is a rich substructure within the problem space, consisting of a range of distinct subproblems that agents must solve to perform well.

Our agents have evolved distinct behavioral responses to the variety of subproblems, as shown by the heterogeneous pattern of strengths and weaknesses. Our results to date indicate that agents with more complex control architectures are easier to evolve robust behavior with, but that this is possible even with only 2 interneurons. We have also found that the relative frequency with which agents are exposed to a given subproblem during evolution influences the likelihood that the agent will develop an appropriate response to it, and manipulated this effect to increase the reliability with which we can evolve agents whose behavior generalizes.

---

[2] Additional trials are actually introduced at generation 601, 1202, 1803… This causes only 14 trials to be added over 9000 generations.

We believe that understanding the dynamics of an evolved agent requires us to also understand the subtleties of the task for which they are evolved. The selective attention task we are using turns out to have a more complex structure than is at first apparent, and analyzing this has helped us to characterize the agents.

There are several directions for future work that we believe will be productive. Our decomposition of the problem space is not yet exhaustive, and extending this will probably allow us to make finer-grained distinctions between behavior patterns. This in turn would support a more detailed analysis of individual agents, linking the internal dynamics of the CTRNNs to their observed behavior. Such an analysis would help us to understand how an agent with as few as 2 interneurons can perform well on subproblems requiring a combination of short-term memory and selective attention.

We have made various tentative claims about the relative difficulty of evolving robust behavior with different numbers of interneurons. Evaluating these claims fully will require a more detailed characterization of the fitness space, which in turn will allow us to further refine the shaping protocol and produce robust behavior more reliably.

## Acknowledgments

This work was supported in part by grant EIA-0130773 from the NSF.